\documentclass[nocompress]{spie}  

\usepackage{amsmath,amsfonts,amssymb}
\usepackage{graphicx}
\usepackage[colorlinks=true, allcolors=blue]{hyperref}
\usepackage{tabularx}
\usepackage{multirow}
\usepackage{multicol}
\usepackage{wrapfig}
\usepackage{makecell}
\usepackage{booktabs}
\usepackage{algorithm}
\usepackage{subcaption}
\usepackage{algpseudocode}
\usepackage{url}
\usepackage{orcidlink}
\usepackage[table]{xcolor}
\newcommand{\topone}[1]{\cellcolor{cyan!5!white}{\textbf{#1}}}
\newcommand{\toptwo}[1]{\cellcolor{orange!8!white}{\underline{#1}}}

\newcolumntype{Y}{>{\centering\arraybackslash}X}

\title{TreeSoc: Tree-Structured Dynamic Reasoning and Tool Synergy for Soccer Video Understanding}

\author[1,2]{Thanh-Nhan Vo}
\author[1,2]{Thanh-Khoi Nguyen}
\author[1,2]{Trong-Thuan Nguyen}
\author[1,2]{\\ Trung-Hoang Le}
\author[1,2]{Minh-Triet Tran}
\affil[1]{University of Science, VNU-HCM, Ho Chi Minh City, Vietnam}
\affil[2]{Vietnam National University, Ho Chi Minh City, Vietnam}

\makeatletter
\renewcommand{\cite}{\@ifnextchar[{\CITE@with}{\CITE@without}}
\def\CITE@without#1{[\citen{#1}]}
\def\CITE@with[#1]#2{[\citen{#2}, #1]}
\makeatother

\begin{document} 
\maketitle

\begin{abstract}
Automated understanding of complex soccer scenarios from video remains a significant challenge for contemporary vision-language models (VLMs), which suffer from shallow cross-modal alignment and exhibit fundamental limitations in multi-step reasoning and coordinated tool integration. We present TreeSoc, a structured reasoning framework that reformulates soccer video question answering as a hierarchical search problem rather than a single-pass prediction. Specifically, TreeSoc employs a dynamic depth-first search (DFS) mechanism that decomposes complex queries into sequentially ordered sub-tasks, enabling iterative reasoning refinement through explicit intermediate states. This tree-structured decomposition naturally supports adaptive tool routing, wherein domain-specific modules are selectively activated and their outputs incorporated at each reasoning node to produce contextually grounded predictions. On SoccerBench, TreeSoc achieves state-of-the-art performance, with accuracies of 85.2\%, 87.4\%, and 82.2\% on TextQA, ImageQA, and VideoQA, respectively. Additionally, TreeSoc further demonstrates strong cross-domain generalization, attaining 74.16\% accuracy on NExT-QA. These results establish structured, tool-augmented tree reasoning as an effective paradigm for robust video understanding. Code is available at: \url{https://github.com/thanhnhan29/TreeSoc}.
\end{abstract}


\section{INTRODUCTION}
\label{sec:Introduction}

Visual Question Answering (VQA) has emerged as an important paradigm for extracting semantic and contextual information from visual data, particularly in dynamic video environments~\cite{wu2017visual}. In the sports domain, especially soccer, VQA supports a wide range of specialized analytical tasks, including foul recognition, player tracking, tactical interpretation, and commentary generation~\cite{rao2025multi,Held2024XVARS-arxiv,rao2024unisoccer}. Beyond answering factual questions, high-level soccer VQA requires models to reason over complex spatio-temporal events, such as tactical buildups, player interactions, and the causal factors underlying refereeing decisions~\cite{Held2024XVARS-arxiv}. These characteristics make soccer video understanding a challenging yet valuable foundation for studying domain-specific visual reasoning.

Despite significant progress in video understanding and sport analysis, existing vision-language models (VLMs) still face notable limitations when applied to complex sports VQA tasks. In particular, most current models rely on single-pass inference and implicit visual-textual alignment, which limits their capacity to construct explicit reasoning trajectories, coordinate specialized perception tools, and revise intermediate decisions~\cite{campbell2024_vlm_limits,rahman2026_vlm_survey}. These limitations are particularly critical in soccer videos, where relevant evidence is often distributed across multiple video frames, grounded in domain-specific concepts, and dependent on external knowledge sources or specialized analytical modules~\cite{domainadapt2025_soccer_vlm,hvu2019_holistic}. Therefore, early visual misinterpretations may propagate directly to the final answer when the reasoning process lacks structured verification and adaptive replanning mechanisms~\cite{moonlight_2026_error_propagation}.

In this paper, we propose TreeSoc, a structured reasoning framework for visual question answering over soccer videos. Rather than treating VQA as a single-pass prediction problem, TreeSoc reformulates it as a multi-step state-space search over a hierarchical reasoning tree~\cite{yin2025toolvqa}. The framework employs a Multimodal Large Language Model (MLLM) as a central coordinator to recursively decompose complex questions into manageable subtasks. For each subtask, the coordinator adaptively selects and invokes domain-expert modules, including YOLO26~\cite{sapkota2025yolo26}, PRTREID~\cite{Mansourian2023Multitask}, and UniSoccer~\cite{rao2024unisoccer}, or retrieves relevant information from external databases~\cite{rao2025multi,mkhallati2023soccernet}. Through a dynamic depth-first search (DFS), TreeSoc updates its execution queue based on intermediate observations, enabling adaptive replanning and reducing the impact of error propagation during multi-step reasoning.

The main contributions of this paper are summarized as follows. First, we introduce a tree-structured reasoning formulation for soccer VQA that explicitly decomposes complex questions into hierarchical subtasks. Second, we design an LLM-driven coordination mechanism that dynamically routes subtasks to specialized perception tools and external knowledge sources. Third, we incorporate adaptive replanning through DFS-based execution, allowing the system to revise its reasoning path according to intermediate results. Finally, we demonstrate the effectiveness of TreeSoc on both domain-specific and general video QA benchmarks. On SoccerBench~\cite{rao2025multi}, TreeSoc achieves 85.2\% on TextQA, 87.4\% on ImageQA, and 82.2\% on VideoQA. Moreover, it shows strong zero-shot generalization on the in-the-wild NExT-QA dataset~\cite{xiao2021next}, reaching 74.16\% accuracy.


\section{RELATED WORK}
\label{sec:relatedwork}

\paragraph{Vision-Language Models for Sports Video Understanding}
Recent advances in sports video understanding have positioned it as a challenging domain that requires substantial domain-specific knowledge for tasks such as foul recognition~\cite{held2023vars,Held2024XVARS-arxiv}, player tracking~\cite{scott2025soccertrack,cioppa2022soccernet}, and commentary generation~\cite{rao2024unisoccer,mkhallati2023soccernet}. Traditionally, these tasks have been addressed using bespoke models optimized for individual objectives. With the emergence of VLMs~\cite{rao2025multi,rao2024unisoccer,Mansourian2023Multitask}, however, research has increasingly shifted toward more unified and holistic formulations of sports video understanding. Nevertheless, existing VLMs remain limited in their ability to coordinate multiple analytical capabilities in dynamic environments such as soccer. This limitation is particularly evident in high-level VQA scenarios, where models must analyze tactical buildups, infer causal game events, and reason over complex spatio-temporal relations. Current VLMs largely rely on implicit visual-textual alignment and therefore lack explicit reasoning mechanisms for adaptively integrating diverse analytical skills.
\vspace{-5mm}
\paragraph{Soccer Video Understanding}
Soccer video understanding has transitioned from primitive event detection to fine-grained spatio-temporal game state reconstruction. Early methodologies primarily focused on isolated tasks such as shot boundary detection or macro-event spotting. However, the introduction of comprehensive datasets like SoccerNet~\cite{cioppa2022soccernet} has catalyzed a paradigm shift toward detailed athlete tracking, team affiliation clustering, and complex, dense video captioning. Recent specialized architectures, such as UniSoccer~\cite{rao2024unisoccer}, attempt to establish universal foundation models for soccer by learning broad semantic representations directly from match broadcasts. Concurrently, automated decision-making frameworks such as VARS~\cite{held2023vars} and X-VARS~\cite{Held2024XVARS-arxiv} have introduced multi-view explainability into refereeing and foul-severity recognition. To systematically evaluate these multi-faceted analytical capabilities, the SoccerBench~\cite{rao2025multi} benchmark was introduced, establishing a rigorous multi-task public test that evaluates models across diverse modalities, including TextQA, ImageQA, and VideoQA. Despite these advancements, existing approaches either rely on rigid, non-adaptive pipelines or expect monolithic models to zero-shot complex tactical queries, highlighting a critical gap in dynamic, multi-step reasoning.

\section{PROBLEM FORMULATION}\label{sec:form}
Given a video $\mathcal{V}$ and a natural language query $q_0$, the objective of specialized sports video question answering is to derive the optimal global solution $\mathcal{A}$. We define the contextual state at execution step $t$ as $S_t$, initialized as $S_0 = \{\mathcal{V}, q_0\}$. Instead of formulating this mapping as a conventional single-pass end-to-end prediction task, we model the problem as a multi-step state-space search over a hierarchical reasoning tree $\mathcal{T} = (\mathcal{V}_{\mathcal{T}}, \mathcal{E}_{\mathcal{T}})$. Let $D_{max}$ and $W_{max}$ denote the maximum allowable search depth and branching width of $\mathcal{T}$, respectively. For any arbitrary sub-task query $q_n \in \mathcal{V}_{\mathcal{T}}$ located at depth $d_n$, its mathematical execution policy $E(q_n, d_n, S_t)$ is conditionally formulated based on an atomic termination constraint $c_n$, which is formally defined in Eqn.~\eqref{eqn:c_n}.

\begin{equation}
E(q_n, d_n, S_t) = \begin{cases} Solve(q_n, S_t), & \text{if } c_n \\ \mathcal{A}_{LLM}(R_n), & \text{otherwise.} \end{cases}
\label{eqn:c_n}
\end{equation}
where $c_n = IsSimple(q_n, S_t) \lor (d_n = D_{max})$, and $Solve(q_n, S_t)$ represents a direct deterministic execution mapping to a specific domain-expert perception module or a local database. In cases where $c_n = \text{False}$, the sub-task space expands dynamically into a sequential execution queue $Q_n^{(0)} = [q_{n,1}, \dots, q_{n,k}]$, where $k \le W_{max}$. The objective of this formulation is to recursively resolve the set of intermediate sub-task outputs $R_n = \{r_{n,1}, \dots, r_{n,m}\}$ to enable the target aggregation function $\mathcal{A}_{LLM}(R_n)$ to compute the final validated state response $\mathcal{A}$.

We consider a diverse set of query types following the SoccerBench taxonomy~\cite{rao2025multi} and the additional tasks in the current evaluation protocol. They are organized into three multimodal clusters. \textit{Text-Based Knowledge and Match Queries} cover non-visual factual reasoning, including background knowledge about players, teams, referees, and venues (Q1), as well as match-specific information such as lineups, coaches, goals, and cards from major European leagues between 2014 and 2024 (Q2). \textit{Static Image Perception Tasks} focus on extracting evidence from individual broadcast frames, including camera status, visual knowledge, jersey numbers, scoreboards, timers, and constrained player counting (Q3-Q7). \textit{Dynamic Video Analysis Tasks} require spatio-temporal reasoning over clips, covering camera transitions, replay grounding, action classification, commentary-related tasks, jersey color reasoning, and multi-view foul recognition with type and severity estimation (Q8-Q14).

\section{THE PROPOSED APPROACH}
\label{sec:method}
\begin{wrapfigure}{r}{0.50\linewidth}
    \centering
    \vspace{-10pt}
    \includegraphics[width=\linewidth]{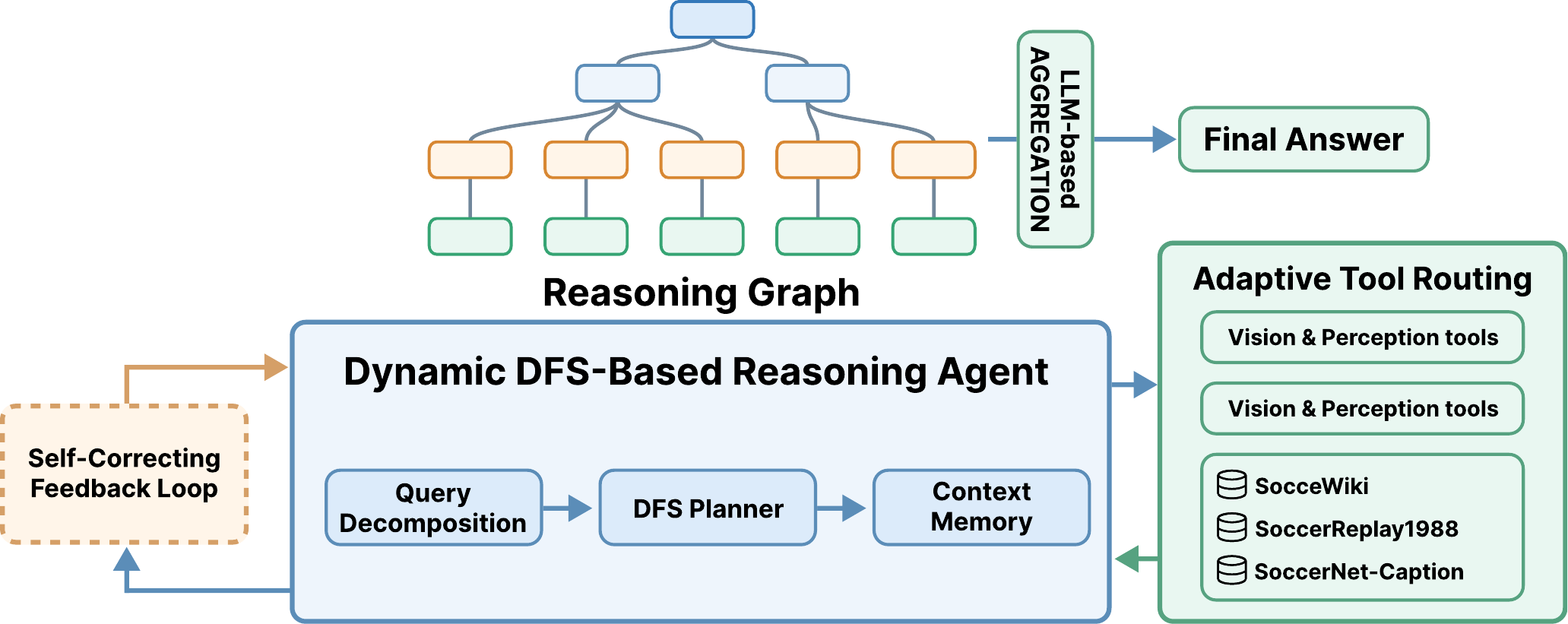}
    \caption{Overview of TreeSoc. TreeSoc decomposes an input query into a DFS-based subtask tree and adaptively invokes domain-specific tools or retrieval sources to update the reasoning context.}
    \label{fig:pipeline}
    \vspace{-15pt}
\end{wrapfigure}

As shown in Figure \ref{fig:pipeline}, TreeSoc reformulates specialized sports visual question answering from a conventional single-pass prediction into a multi-step state-space search over a hierarchical reasoning tree. 
Section \ref{sec:Dynamic DFS-based Reasoning Agent} details the core of TreeSoc, where a Large Language Model operates as a dynamic depth-first search (DFS) agent that recursively decomposes complex, non-trivial input queries into a structured subtask tree
. Subsequently, Section \ref{sec:Query-Specific Strategies and Tool Synergy} outlines the query-specific strategies and tool synergy, explaining how the central agent adaptively routes and triggers domain-specific perception tools (such as YOLO26\cite{sapkota2025yolo26} and UniSoccer\cite{rao2024unisoccer}) or queries external retrieval databases (like SoccerWiki\cite{rao2025multi} and SR-1988\cite{rao2024unisoccer}) via explicit tool calls. By continuously utilizing these intermediate results to dynamically update the context, TreeSoc establishes a self-correcting feedback loop that coordinates multifaceted analytical skills and mitigates error propagation before aggregating the response.

\subsection{Dynamic DFS-based Reasoning Agent}

\label{sec:Dynamic DFS-based Reasoning Agent}
The core of TreeSoc is a recursive depth-first reasoning agent that decomposes complex queries, invokes specialized tools, and updates its plan based on intermediate evidence. Unlike static pipelines with predefined execution orders, TreeSoc incrementally constructs a query-specific reasoning tree, which is well-suited for soccer understanding, where later steps often depend on earlier visual, temporal, or knowledge-based observations. Given an input $x$ and an initial query $q_0$, TreeSoc builds a reasoning tree $\mathcal{G}_x=(\mathcal{V}_x,\mathcal{E}_x)$, where each node is a sub-query and each directed edge represents a dependency. The root corresponds to $q_0$ intermediate nodes, which require further decomposition, and leaf nodes are atomic operations resolved by a tool call or direct inference. The search is bounded by $D_{\max}$ and branching width $W_{\max}$. Alg.~\ref{alg:dfs_core} summarizes the recursive inference process.

\begin{algorithm}[!t]
\scriptsize
\small\caption{Core DFS-Based Reasoning}
\label{alg:dfs_core}
\begin{algorithmic}[1]
\Require Context $ctx$, evidence state $S$, tool bank $\mathcal{T}$, depth $d$
\Ensure Response $r$
\Procedure{DFS-Reason}{$ctx, S, d$}
    \State \textbf{if } $d \geq D_{\max}$ \textbf{ then return } \Call{Direct-Answer}{$ctx,S$}
    \State $(m,\mathcal{Q}) \gets \Call{Think}{ctx,d}$ \Comment{Decision and initial subtask queue}
    \State \textbf{if } $m=\textsc{Direct}$ \textbf{ or } $\mathcal{Q}=\emptyset$ \textbf{ then return } \Call{Direct-Answer}{$ctx,S$}
    \While{$\mathcal{Q}\neq\emptyset$}
        \State $q_i \gets \Call{Pop}{\mathcal{Q}}$ \Comment{Get next subtask}
        \State $r_i \gets 
        \begin{cases}
        \Call{Execute-Task}{q_i,ctx,S,\mathcal{T}}, & \Call{Is-Leaf-Task}{q_i}\\
        \Call{DFS-Reason}{q_i,S,d+1}, & \text{otherwise}
        \end{cases}$ \Comment{Recursive reasoning}
        \State $S \gets S \cup \{(q_i,r_i)\}$
        \State $a \gets \Call{Reflect}{q_i,r_i,S}$ \Comment{Evaluate progress}
        \State \textbf{if } $a=\textsc{Stop}$ \textbf{ then break}
        \State $\mathcal{Q} \gets \Call{Update-Queue}{\mathcal{Q},a}$ \Comment{Insert, Modify, Skip, or Rerun}
    \EndWhile
    \State \Return \Call{Synthesize}{$ctx,S$}
\EndProcedure
\end{algorithmic}
\end{algorithm}


    

        
        
        


\vspace{-3mm}
\paragraph{Node planning and termination.}
At each node $n$, the coordinator invokes \textsc{Think} to determine whether the current text query can be answered directly or should be decomposed into an ordered subtask queue $\mathcal{Q}_n=[q_{n,1},q_{n,2},\ldots,q_{n,k}], \qquad k\leq W_{\max}.$ In particular, this ordering captures dependencies among the sub-tasks. For example, a player-related knowledge query may first require action localization, followed by player identification and an external database lookup. If the coordinator selects the \textsc{Direct} mode, produces no valid sub-tasks, or reaches $D_{\max}$ value, the node is resolved using direct inference without further expansion.


\vspace{-3mm}
\paragraph{Recursive depth-first traversal.}
For each subtask $q_{n,i}$, TreeSoc first determines whether it is a leaf node. Leaf nodes are atomic operations that can be resolved by a specialized tool or direct inference. In particular, non-leaf reasoning nodes instead construct a child context $C_i$ and recursively invoke \textsc{DFS-Reason} at depth $d+1$. Then, the recursive call is completed before the agent continues with $q_{n,i+1}$. Consequently, TreeSoc fully explores the current reasoning branch before returning to the remaining sibling subtasks, which yields a depth-first traversal of the dynamically constructed tree. This recursive execution more accurately reflects the implemented inference procedure than a breadth-first expansion or a globally predefined plan.

\vspace{-3mm}
\paragraph{Tool execution and evidence accumulation.}
For an atomic sub-task, the coordinator selects a suitable tool from the tool bank $\mathcal{T} = \{\tau_k\}_{k=1}^K$ through a routing function conditioned on the subtask and the accumulated evidence. In particular, the selected tool may correspond to a visual perception model, temporal analysis module, domain-specific recognizer, or external retrieval system. In addition, its output is appended to the evidence state $S_{t+1} = S_t \cup \{(q_{n,i}, r_{n,i})\}$. When a subtask depends on an earlier result, \textsc{Resolve-Dependencies} replaces its unresolved references with the corresponding evidence in the state $S_t$ of the current $t$. As a result, this allows later operations to use concrete outputs, such as a recognized player identity, localized temporal segment, or retrieved match record, instead of relying on assumptions made during the initial decomposition.

\vspace{-3mm}
\paragraph{Reflection, replanning, and synthesis.}
After each subtask is resolved, TreeSoc updates the remaining execution plan according to the newly collected evidence $\left( \mathcal{Q}_n^{(i+1)}, a_i \right) =
    \Phi_{\mathrm{LLM}} \left( \mathcal{Q}_n^{(i)}, q_{n,i}, r_{n,i}, S_{t+1} \right),
$
where $a_i \in \{ \text{\textsc{Keep}}, \text{\textsc{Modify}}, \text{\textsc{Insert}}, \text{\textsc{Skip}}, \text{\textsc{Retry}}, \text{\textsc{Stop}} \}$.
This reflection step allows the coordinator to revise, skip, rerun, or terminate sub-tasks based on intermediate observations, so reliable evidence can simplify the remaining plan while ambiguous evidence can trigger verification. When the sub-task queue is exhausted or \textsc{Stop} is selected, the collected evidence is aggregated into the current node response, $r_n = \mathcal{A}_{\mathrm{LLM}}(C_n, S)$. The response is returned to the parent node as new evidence, and the same aggregation process at the root produces the multiple-choice outputs constrained to the provided options and open-ended outputs following the required format.

\subsection{Query-Specific Strategies and Tool Synergy}
\label{sec:Query-Specific Strategies and Tool Synergy}

\paragraph{Visual Grounding and Player-centric Perception (Q4-Q7, Q9, Q10):} To reduce alignment noise in standard vision-language models, TreeSoc uses specialized perception modules for visual grounding, temporal localization, and player-centric reasoning. For image- and video-based queries, the agent invokes task-specific tools such as face recognition for player identification (Q4), jersey-region extraction from sparsely sampled frames for number recognition (Q5), OCR over scoreboard regions for score and time QA (Q6), replay retrieval through spatio-temporal similarity matching (Q9), and UniSoccer~\cite{rao2024unisoccer} for action classification (Q10). For player counting and identification (Q7), TreeSoc further decomposes the task into detection, re-identification, team clustering, and tactical assignment. Specifically, YOLO26 first extracts candidate player boxes, which are filtered by confidence and passed to a re-identification model to obtain player embeddings. These embeddings are clustered to infer team affiliation, while a tactical mapping function uses pitch geometry and spatial distribution to assign attacking and defending roles. The resulting identities, timestamps, action labels, and team-level assignments are injected into the reasoning context as compact semantic evidence for downstream reasoning.
\vspace{-3mm}
\paragraph{Retrieval-Augmented Reasoning (Q11, Q12, Q13):}
To resolve language-grounded queries and mitigate the severe logical hallucinations common in unconstrained video-to-text generation, we formalize a Retrieval-Augmented Generation (RAG) framework. Given a target natural language query $q_n$, a structured suite of semantic retrieval operators $\mathcal{K} = \{\text{MatchSearch}, \text{EntitySearch}\}$ is invoked to query external domain-specific knowledge bases $\mathcal{D}_{\text{ext}} = \{\mathcal{D}_{\text{Replay}}, \mathcal{D}_{\text{Caption}}\}$, fetching the top-$k$ relevant factual contexts $\mathcal{R}_{\text{ctx}} = \text{Top-}k \left( \bigcup_{k \in \mathcal{K}} k(q_n \mid \mathcal{D}_{\text{ext}}) \right).$ The retrieved text blocks, encompassing fine-grained match statistics such as team formations, precise lineups, and referee data, are dynamically appended to the current reasoning state via a set union operator, establishing an updated context formulation $\mathcal{S}_{t+1} = \mathcal{S}_t \cup \mathcal{R}_{\text{ctx}}$. Therefore, this context accumulation explicitly grounds the coordinator LLM's deductive trajectory. It is worth noting that for Jersey Color Relevant QA ($Q_{13}$) , our preliminary empirical analysis revealed substantial label noise within the public test bed annotations.

\vspace{-3mm}
\paragraph{Foul and Camera Understanding (Q3, Q8, Q14):}
For fine-grained foul recognition and camera-related tasks, TreeSoc relies on specialized perception modules to produce compact semantic evidence for the reasoning agent. For Q14, a multi-view spatial-temporal model estimates foul categories and severity levels, which are aggregated across views using voting or mode-based consensus before being injected into the context memory. For Q3 and Q8, a camera classifier predicts broadcast camera status from frame-level visual features, while shot-level segmentation allows the model to track camera transitions. These structured labels provide high-level visual evidence, enabling the coordinator to reason over foul severity, camera status, and viewpoint changes.

\section{EXPERIMENT RESULTS}
\label{sec:experiments}
\subsection{Experimental Setup}
\begin{table*}[!t]
\centering
\small
\setlength{\tabcolsep}{4pt}
\renewcommand{\arraystretch}{1.15}
\caption{Quantitative comparison (\%) on SoccerBench test set against APIs and open-source baselines reported from the official leaderboard. TreeSoc achieves the best overall performance across TextQA, ImageQA, and VideoQA categories. Best results in \textbf{bold} and second-best results are \toptwo{underlined}; "–" is unreported results.}
\resizebox{\textwidth}{!}{%
\begin{tabular}{l|cc|ccccc|ccccccc|ccc}
\toprule
\multirow{2}{*}{Model} & \multicolumn{2}{c|}{TextQA} & \multicolumn{5}{c|}{ImageQA} & \multicolumn{7}{c|}{VideoQA} & \multicolumn{3}{c}{Overall} \\
& Q1 & Q2 & Q3 & Q4 & Q5 & Q6 & Q7 & Q8 & Q9 & Q10 & Q11 & Q12 & Q13 & Q14 & Text & Image & Video \\
\midrule
\multicolumn{18}{c}{\textit{Commercial APIs}} \\
\midrule
Claude 3.7 Sonnet & 58.1 & 58.2 & 51.3 & 32.0 & 63.3 & 63.9 & - & 39.8 & 26.8 & 48.3 & 49.3 & 38.6 & 43.9 & 45.5 & 58.1 & 47.1 & 43.4 \\
Gemini 2.0 Flash & 61.9 & 52.2 & 63.2 & 41.0 & 88.5 & 67.3 & - & 59.0 & \toptwo{46.0} & 56.1 & 62.7 & 42.8 & \toptwo{52.4} & 55.0 & 57.6 & 56.5 & 54.0 \\
GPT-4o & 64.0 & 58.5 & \toptwo{76.7} & 46.0 & \toptwo{89.6} & 70.6 & - & 61.3 & 40.0 & 66.4 & 70.0 & 43.7 & 49.9 & \toptwo{59.7} & 61.6 & 62.3 & 57.5 \\
\midrule
\multicolumn{18}{c}{\textit{Open-Source Models}} \\
\midrule
DeepSeek-v3 & 56.0 & 49.5 & - & - & - & - & - & - & - & - & - & - & - & - & 53.1 & - & - \\
DeepSeek-R1 & 68.3 & 51.1 & - & - & - & - & - & - & - & - & - & - & - & - & 60.6 & - & - \\
Qwen2.5-VL (7B) & 35.6 & 53.5 & 58.5 & 35.8 & 82.0 & 66.0 & - & 56.8 & 31.6 & 52.2 & 51.6 & 35.0 & 46.9 & 50.7 & 43.6 & 52.4 & 46.8 \\
Qwen2.5-VL (72B) & 49.4 & 37.7 & 66.5 & 45.9 & 87.0 & 67.5 & - & \toptwo{67.5} & 19.5 & 58.8 & 58.5 & 51.0 & 49.0 & 58.7 & 44.2 & 59.3 & 53.2 \\
LLaVA-onevision (7B) & 37.4 & 42.5 & 47.6 & 32.3 & 84.5 & 62.8 & - & 38.2 & 23.0 & 24.5 & 26.8 & 35.5 & 29.1 & 49.3 & 39.6 & 48.1 & 30.3 \\
VideoLLaMA3 (7B) & - & - & 54.3 & 41.9 & 78.6 & 66.3 & - & 49.5 & 23.3 & 39.6 & 43.6 & 35.0 & 46.3 & 43.0 & - & 50.4 & 40.4 \\
LLaVA-Video (7B) & - & - & 59.3 & 39.6 & 38.0 & 61.0 & - & 50.9 & 26.3 & 41.2 & 49.8 & 41.8 & 48.4 & 59.3 & - & 54.1 & 45.0 \\
VideoChat-Flash-Qwen2 (7B) & - & - & - & - & - & - & - & 51.8 & 21.9 & 40.5 & 48.7 & \toptwo{54.8} & 42.2 & 48.3 & - & - & 45.0 \\
\midrule
SoccerAgent (MCQ) & \topone{95.9} & \toptwo{71.4} & 73.4 & \toptwo{69.2} & 85.7 & \toptwo{75.8} & - & 51.1 & 35.7 & \topone{85.0} & \toptwo{72.9} & 49.0 & 46.0 & 55.5 & \toptwo{85.0} & \toptwo{73.3} & \toptwo{60.9} \\
\textbf{TreeSoc} (Ours) & \toptwo{88.0} & \topone{78.0} & \topone{90.0} & \topone{90.0} & \topone{100.0} & \topone{87.0} & \topone{76.0} & \topone{90.0} & \topone{70.0} & \toptwo{84.0} & \topone{100.0} & \topone{90.0} & \topone{54.0} & \topone{67.0} & \topone{85.2} & \topone{87.4} & \topone{82.2} \\
\bottomrule
\end{tabular}%
}\label{tab:1}
\end{table*}
\paragraph{Implementation Details.}
We employ Qwen3.5-9B as the core LLM, deployed via the vLLM~\cite{kwon2023efficient} engine to ensure high-throughput inference. Auxiliary perception modules are integrated as callable tools. Specifically, we utilize a domain-finetuned YOLO26~\cite{sapkota2025yolo26} object detector configured with a confidence threshold of $0.3$, alongside PRTReID~\cite{Mansourian2023Multitask} for player re-identification. Furthermore, we incorporate a dedicated face recognition module, which is setup on an aggregated image dataset curated from SoccerWiki~\cite{rao2025multi} and \textit{UniSoccer}~\cite{rao2024unisoccer} to enhance domain-specific accuracy. The generation parameters are configured as follows: temperature $0.7$, Top-$p$ $0.8$, Top-$k$ $20$, presence penalty $1.5$, and repetition penalty $1.0$. For reproducibility, the random seed is fixed to $0$.
\vspace{-6mm}
\paragraph{Datasets and Metrics.}
We evaluate TreeSoc on SoccerBench~\cite{rao2025multi} and NExT-QA~\cite{xiao2021next}. SoccerBench contains 500 public test samples for domain-specific multi-task soccer understanding, while NExT-QA includes 8,564 samples for evaluating temporal localization and causal reasoning in in-the-wild videos. For SoccerBench~\cite{rao2025multi}, we use accuracy as the primary metric and report results across both individual subtasks and three modality-specific groups: TextQA, ImageQA, and VideoQA. For NExT-QA~\cite{xiao2021next}, we evaluate on the official test set using overall accuracy and further report performance across causal, descriptive, and temporal reasoning categories.


\subsection{Experimental Discussion}
\begin{wrapfigure}{r}{0.50\linewidth}
    \centering
    \vspace{-15pt}
    \includegraphics[width=\linewidth]{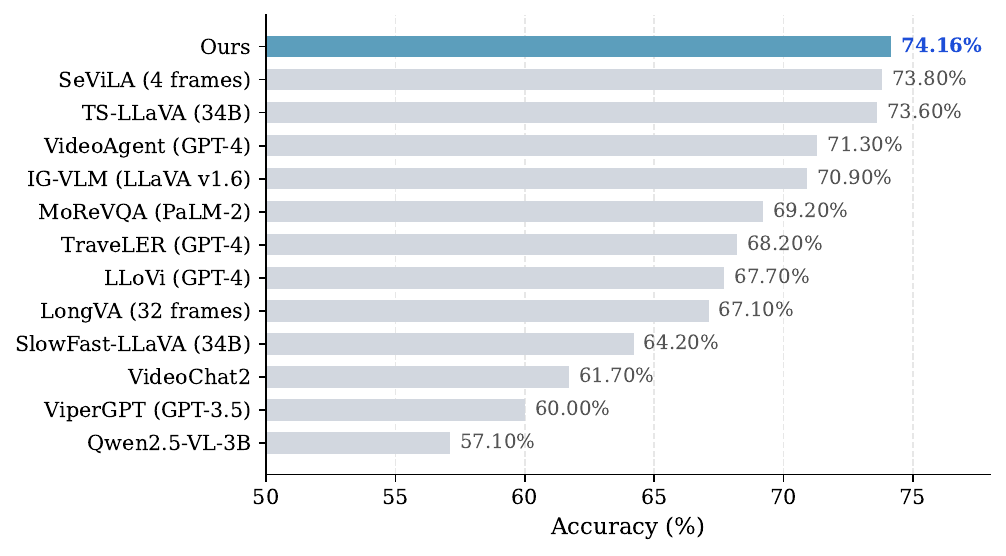}
    \caption{Evaluation on the NExT-QA dataset.}
    \label{fig:NExT-QA}
    \vspace{-12pt}
\end{wrapfigure}

\paragraph{Results on SoccerBench.}
Table~\ref{tab:1} reports the results on the SoccerBench~\cite{rao2025multi} public test. TreeSoc achieves strong performance across modalities, obtaining 85.2\% text accuracy, 87.4\% image accuracy, and 82.2\% video accuracy. It also shows clear advantages over open-source multimodal models, including Qwen2.5-VL (72B) and LLAVA-Video~\cite{zhang2024llava} (7B). Notably, on commentary generation (Q11), TreeSoc reaches 100.0\%, substantially outperforming Qwen2.5-VL (72B), which achieves 58.5\%, suggesting that coordinating specialized perception tools through structured tree search can be more effective than relying solely on larger multimodal backbones. More importantly, TreeSoc improves upon SoccerAgent~\cite{rao2025multi}, a soccer-specific agentic baseline. Unlike SoccerAgent, which relies on relatively fixed tool-calling or multi-agent prompting, TreeSoc formulates reasoning as a dynamic task-decomposition tree $\mathcal{T}=(\mathcal{V},\mathcal{E})$, allowing the agent to update its remaining subtask queue as new intermediate evidence $S_{t+1}$ is obtained. This adaptive mechanism helps reduce error propagation and leads to better visual-temporal reasoning, as shown by the perfect score on Q5 compared with 85.7\% for SoccerAgent, and stronger fine-grained foul recognition on Q14, where TreeSoc achieves 67.0\% versus 55.5\%. As a result, these results demonstrate the effectiveness of dynamic tree-structured reasoning for soccer video understanding.

\vspace{-6mm}
\paragraph{Results on in-the-wild Videos.}

To evaluate the robustness and generalization of our approach beyond soccer-specific settings, we further test it on the official NExT-QA~\cite{xiao2021next} test set, a widely used benchmark for temporal understanding and causal reasoning in unconstrained real-world videos. Compared with SoccerBench, NExT-QA covers more diverse everyday activities, objects, and interactions, making it suitable for assessing cross-domain transfer. As shown in Figure~\ref{fig:NExT-QA}, TreeSoc achieves an overall accuracy of 74.16\%, slightly outperforming SeViLA at 73.80\%, and surpassing TS-LLaVA~\cite{qu2024tsllava}, VideoAgent~\cite{wang2024videoagent}, and MoReVQA~\cite{min2024morevqa} by 0.56, 2.86, and 4.96 percentage points, respectively. These results suggest that TreeSoc’s effectiveness is not solely tied to soccer-specific tools or knowledge; rather, its dynamic task decomposition, intermediate context accumulation, and adaptive reasoning strategy can transfer to broader video question answering tasks involving causal, temporal, and descriptive reasoning. The consistent performance across SoccerBench~\cite{rao2025multi} and NExT-QA~\cite{xiao2021next} highlights the potential of tree-structured, tool-augmented reasoning for general video understanding.

\vspace{-6mm}
\paragraph{Fine-grained Reasoning Analysis.} To investigate the individual contributions of our dynamic tree-structured reasoning, we conduct an ablation analysis on the NExT-QA~\cite{xiao2021next} test set, in which TreeSoc achieves a robust Average accuracy of 74.16\%. As summarized in Table~\ref{tab:nextqa_qtype}, in causal reasoning, TreeSoc scores 75.64\% on How and 73.49\% on Why tasks. While monolithic models struggle to link non-adjacent factors in long videos, TreeSoc isolates causal conditions by decomposing queries into a hierarchical sub-task queue $Q_n^{(0)}$ , where sequential context accumulation ($S_{t+1} = S_t \cup \{r_{n,i}\}$) provides solid anchors that prevent logical hallucinations.  For Descriptive Reasoning, our proposed approach achieves 67.70\% on Count, an exceptional peak of 90.87\% on Location, and 81.97\% on Other.  Finally, TreeSoc scores 74.29\% on Present and 67.27\% on Prev\&Next tasks. 

\begin{table*}[!h]
\centering
\caption{Detailed performance (\%) breakdown of our proposed approach, TreeSoc, across different question categories on the NExT-QA test set, including causal, descriptive, and temporal reasoning tasks.}
\label{tab:nextqa_qtype}
\small
\setlength{\tabcolsep}{10pt} 
\renewcommand{\arraystretch}{1.2}

\begin{tabular}{l|cc|ccc|cc|c} 
\toprule
\multirow{2}{*}{\textbf{Method}} 
& \multicolumn{2}{c|}{Causal}
& \multicolumn{3}{c|}{Descriptive}
& \multicolumn{2}{c|}{Temporal}
& \multirow{2}{*}{\textbf{Avg.}} \\

\cmidrule(lr){2-3} 
\cmidrule(lr){4-6} 
\cmidrule(lr){7-8}

& How & Why & Count & Location & Other & Present & Prev\&Next & \\

\midrule


\rowcolor{blue!5} 
\textbf{TreeSoc} & \textbf{75.64} & \textbf{73.49} & \textbf{67.70} & \textbf{90.87} & \textbf{81.97} & \textbf{74.29} & \textbf{67.27} & \textbf{74.16} \\

\bottomrule
\end{tabular}
\end{table*}

\begin{figure}[!t]
    \centering

    \begin{subfigure}[t]{0.48\linewidth}
        \centering
        \includegraphics[width=\linewidth]{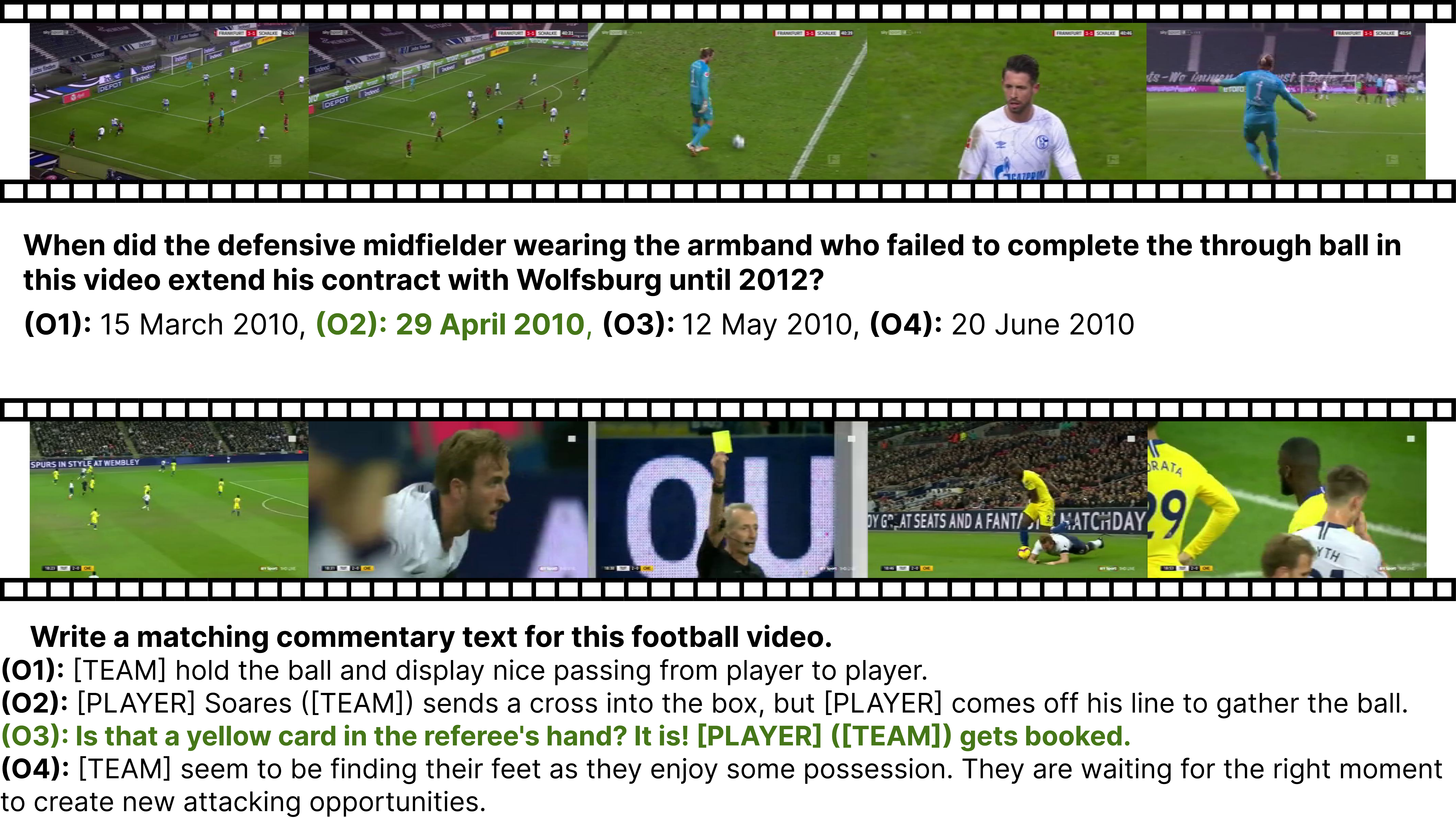}
        \caption{Representative successful predictions by TreeSoc.}
        \label{fig:success}
    \end{subfigure}
    \hfill
    \begin{subfigure}[t]{0.48\linewidth}
        \centering
        \includegraphics[width=\linewidth]{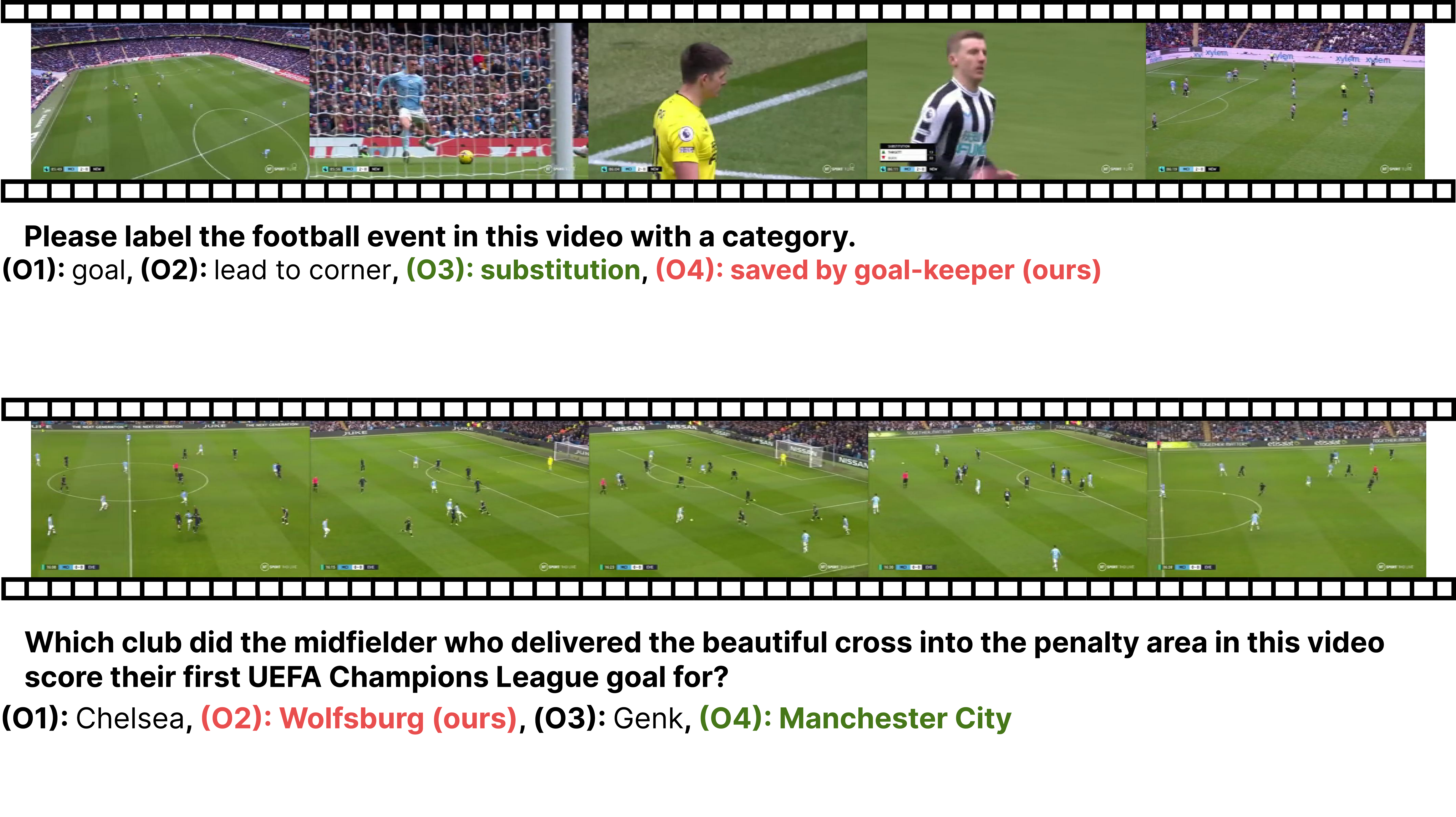}
        \caption{Representative failure cases of TreeSoc.}
        \label{fig:fail}
    \end{subfigure}

    \caption{Qualitative analysis of TreeSoc on soccer video understanding tasks. The examples illustrate both successful predictions and failure cases across diverse reasoning scenarios. \textit{Best view in color and zoom in}.}
    
    \label{fig:qualitative_examples}
\end{figure}

\vspace{-6mm}
\paragraph{Qualitative Results.} Figure~\ref{fig:success} presents representative successful examples across soccer understanding tasks, showing TreeSoc’s ability to handle diverse reasoning requirements, including visual-knowledge reasoning and commentary selection. These cases require temporal modeling, visual grounding, and external knowledge integration: the model grounds the relevant player/action before retrieving structured information for knowledge-based answers, while in commentary selection, it identifies the referee’s booking action and matches it with the most semantically consistent description. Figure~\ref{fig:fail} shows limitations, where TreeSoc may be distracted by visually salient secondary actions, such as predicting a goalkeeper save instead of a substitution, or suffer from error propagation when incorrect player grounding leads to wrong results and a plausible but incorrect prediction.

\section{CONCLUSION}
\label{sec:conclusion}
In this paper, we introduced TreeSoc, a dynamic tree-structured framework for specialized sports video question answering. Instead of relying on monolithic single-pass prediction, TreeSoc uses a large language model as a central coordinator to decompose complex queries into a hierarchical task tree $\mathcal{T}$ and navigate it through dynamic depth-first search. This design enables context propagation, adaptive tool orchestration, and real-time replanning via the transition function $\Phi_{LLM}$, helping reduce visual-temporal noise and error propagation. Experiments on SoccerBench~\cite{rao2025multi} show that TreeSoc achieves competitive performance across TextQA, ImageQA, and VideoQA tasks, outperforming commercial APIs and domain-specific agentic baselines, while zero-shot results on NExT-QA~\cite{xiao2021next} further indicate its generalization potential. Nevertheless, TreeSoc still depends on the reliability of its perception tools, retrieval modules, and intermediate reasoning steps, and its tree-search process introduces additional computational overhead compared with single-pass VLM inference. Future work will focus on more efficient search strategies, uncertainty-aware replanning, and broader adaptation to general video understanding.

\acknowledgments
Trong-Thuan Nguyen was funded by the PhD Scholarship Programme of Vingroup Innovation Foundation (VINIF), VinUniversity, code VINIF.2025.TS63. The authors would like to acknowledge Saigon AI Hub, jointly established by VNG Group and Vietnam National University Ho Chi Minh City, for providing computing infrastructure, resources, and a collaborative research environment that supported this work.

\newpage

\bibliography{main} 
\bibliographystyle{spiebib} 

\end{document}